\documentclass{Interspeech2024}
\usepackage{multirow, url}

\interspeechcameraready

    \title{Speak \& Improve Challenge 2025: Tasks and Baseline Systems}

\name[affiliation={1}]{Mengjie}{Qian}
\name[affiliation={1}]{Kate}{Knill}
\name[affiliation={1}]{Stefano}{Banno}
\name[affiliation={1}]{Siyuan}{Tang}
\name[affiliation={1}]{Penny}{Karanasou}
\name[affiliation={1}]{Mark J.F.}{Gales}
\name[affiliation={2}]{Diane}{Nicholls}

\address{
  $^1$ALTA Institute/MIL Lab, Dept. of Engineering, University of Cambridge, UK\\
  $^2$Cambridge University Press and Assessment, UK
  }
\email{mq227@cam.ac.uk,kmk1001@cam.ac.uk}


\begin{document}

\maketitle

\begin{abstract}
    This paper presents the ``Speak \& Improve Challenge 2025: Spoken Language Assessment and Feedback'' -- a challenge associated with the \href{https://sites.google.com/view/slate-2025}{ISCA SLaTE 2025 Workshop}. The goal of the challenge is to advance research on spoken language assessment and feedback, with tasks associated with both the underlying technology and language learning feedback. Linked with the challenge, the \href{https://englishlanguageitutoring.com/datasets/speak-and-improve-corpus-2025}{Speak \& Improve (S\&I) Corpus 2025} is being pre-released, a dataset of L2 learner English data with holistic scores and language error annotation, collected from open (spontaneous) speaking tests on the \href{https://speakandimprove.com}{Speak \& Improve} learning platform.
    The corpus consists of approximately 315 hours of audio data from second language English learners with holistic scores, and a 55-hour subset with manual transcriptions and error labels. The Challenge has four shared tasks: Automatic Speech Recognition (ASR), Spoken Language Assessment (SLA), Spoken Grammatical Error Correction (SGEC), and Spoken Grammatical Error Correction Feedback (SGECF). Each of these tasks has a closed track where a predetermined set of models and data sources are allowed to be used, and an open track where any public resource may be used. Challenge participants may do one or more of the tasks. This paper describes the challenge, the S\&I Corpus 2025, and the baseline systems released for the Challenge.
    \end{abstract}


\section{Introduction}
A key part of learning any language is gaining the ability to speak the language. As we learn we need to know how good our speaking ability is and get feedback to help us improve. Automated systems can help in this, alongside or in place of teachers, but creating robust and accurate automated tools for spoken language assessment and feedback has significant challenges. Any system must handle natural open, spontaneous, speech where the text is not known in advance, and additionally has non-standard accents, pronunciations and grammar compared to L1 (first language) speakers. A major difficulty for researchers in this area is the shortage of high-quality labelled data which has limited research. 
To tackle these issues, we have launched the Speak \& Improve (S\&I) Challenge 2025, featuring a series of tasks aimed at driving progress in this field.

At the core of the challenge is the release of the Speak \& Improve (S\&I) Corpus 2025~\cite{knill2024sandi}, a rich annotated dataset of second language (L2) learner speech collected from the Speak \& Improve platform. This dataset consists of approximately 340 hours of learner speech from speakers with diverse L1 backgrounds and proficiency levels ranging from A2 to C1 of the Common European Framework of Reference (CEFR)~\cite{cefr2001}. It includes detailed annotations such as transcriptions, disfluency labels, grammatical error corrections, and proficiency scores. This breadth of speaker attributes and annotation types makes the S\&I Corpus a unique resource for developing inclusive and accurate tools for spoken language assessment and feedback.

The Challenge comprises four shared tasks, designed to advance spoken language technology on the one hand and improve automated language learning assessment and feedback on the other. The two primary tasks -- Spoken Language Assessment (SLA) and Spoken Grammatical Error Correction Feedback (SGECF) -- focus on language learning, with an emphasis on assessing learner proficiency and providing corrective feedback. Two supplementary tasks -- Automatic Speech Recognition (ASR) and Spoken Grammatical Error Correction (SGEC) -- focus on the technology side, aiming to improve the tools used in speech processing.  Each task presents its own unique challenges, but together they aim to solve critical problems in building comprehensive and robust automated systems for language education. Participants can choose to do one or more tasks, on a closed or open track. Although similar challenges exist, this challenge is the first-ever challenge focusing on SLA, SGEC and SGECF.

This paper describes the challenge tasks and the baseline systems designed to serve as the starting points for participants.

\section{Related Work}
\label{sec:related}
High-quality datasets are essential for advancing research in spoken language assessment and feedback, yet current publicly available resources are limited in their diversity, scope and annotation types. The Speak \& Improve Corpus 2025 stands out by addressing these gaps, offering a comprehensive dataset with rich annotations and diverse speaker attributes.

Existing datasets are restricted in terms of speakers' L1 background. For instance, the International Corpus Network of Asian Learners of English (ICNALE)~\cite{ishikawa2023icnale} focuses exclusively on Asian L1s and only features CEFR levels from A2 to B2, which are pre-assigned based on the learners' previously obtained language certificates. Similarly, the CLES corpus~\cite{coulange2024corpus} is restricted to L1 French speakers with CEFR levels from B1 to B2. While the LearnerVoice dataset~\cite{kim2024learnervoice} contains annotations on grammatical errors and disfluencies, it is limited to L1 Korean speakers and has not yet been publicly released.
Other datasets focus on narrow aspects of spoken language processing. Speechocean762~\cite{zhang2021speechocean} and L2-ARCTIC~\cite{zhao2018l2arctic} primarily address pronunciation assessment of read speech rather than open-speaking (spontaneous) tasks as in Speak \& Improve Corpus 2025. ETLT Italian-L1 dataset~\cite{gretter2021etlt} is another example of resource limited to a specific L1 group, reducing its generalisability. Meanwhile, datasets such as NICT-JLE\cite{izumi2004nictjle} and KISTEC~\cite{kanzawa2022kistec} support disfluency detection and grammatical error correction but lack respective audio recordings, restricting their application in spoken language research.

The Speak \& Improve Corpus 2025 addresses these limitations by providing recordings and annotated transcriptions from speakers with diverse L1 backgrounds and CEFR proficiency levels (A2-C1). Its annotations cover proficiency scores, disfluencies, and grammatical error corrections, enabling research across a wide range of spoken language processing. This breadth of annotation and speaker diversity makes it a valuable resource for developing accurate and inclusive automated tools for spoken language assessment and feedback.

Shared Tasks and challenges play an important role in driving progress in research by fostering innovation and collaboration. Written grammatical error correction (GEC) has been studied in shared tasks over the past 15 years, including the HOO 2011 Pilot Shared Task~\cite{dale2011hoo}, the CoNLL-2013 Shared Task~\cite{ng2013conll}, the CoNLL2014 Shared Task~\cite{ng2014conll}, the BEA-2019 Shared Task~\cite{bryant2019bea}, and MULTIGEC-2025~\cite{multigec2024}. These tasks have driven advancements in detecting and correcting errors in written texts from learners. However, spoken GEC presents unique challenges not addressed by written GEC tasks, such as handling disfluencies, varied accents, and spontaneous speech patterns, requiring new approaches and innovation. The Speak \& Improve Challenge 2025 is the first to address spoken GEC, making it a significant contribution to the field.

In addition, while there have been other language assessment tasks, they often target specific aspects of assessment. For example, competitions like the Automated Student Assessment Prize (ASAP) 2012~\cite{hammer2012asap} have targeted written essay scoring for L1 learners. The various editions of the Spoken CALL Shared Tasks have focused on grammar and semantic meaning~\cite{baur2017overview,baur2018overview,baur2019overview}. 
The Speak \& Improve Challenge 2025, however, is the first to focus on holistic spoken language assessment, which provides a comprehensive score reflecting the L2 learner's overall proficiency.  

By introducing the Speak \& Improve Corpus 2025 and releasing tasks such as Spoken Language Assessment (SLA), Spoken Grammatical Error Correction (SGEC), and Spoken Grammatical Error Correction Feedback (SGECF), the challenge offers a unique opportunity for people to advance spoken language technology.

\section{Data Resources}
This Section describes the Speak \& Improve Corpus 2025 released for the Speak \& Improve Challenge 2025, and other open-source corpora used in developing the baseline systems for the challenge.

\subsection{Speak \& Improve Corpus 2025}
\label{sec:data_sandi}
The Speak \& Improve (S\&I) Corpus 2025~\cite{knill2024sandi} is an L2 learner speech data designed to advance research in spoken language assessment and feedback. Selected from recordings in the S\&I version 1 platform~\cite{nicholls23_interspeech} from 2019 to 2024, the corpus provides a diverse range of learner audio recordings annotated with manual transcriptions, disfluency annotations, and grammatically corrected transcripts, alongside proficiency scores. Each S\&I test is taken in the same order from five-part practice tests:
\begin{itemize}
    \item Part 1: Interview. Answer questions about yourself. 8 short responses to question prompts. 10 seconds to respond to the first half, 20 seconds for the second. The first two questions are not marked.
    \item Part 2: Read Aloud. 8 sentences to be read aloud.
    \item Part 3: Long Turn 1. Give your opinion. The learner has 1 minute to give their opinion on a specific topic using 3 questions to guide them. 
    \item Part 4: Long Turn 2. Give a presentation about a graphic. The learner has 1 minute to describe a process depicted in a diagram. 
    \item Part 5: Communication Activity. Answer questions about a topic. The learner responds to 5 questions relating to an overall topic, each for up to 20 seconds.
\end{itemize}
The read-aloud Part 2 data will not be released as part of this Challenge to focus on open speaking tasks. A complete test for the purposes of the Challenge is defined as the four open parts (1, 3, 4 and 5).

Annotations in the S\&I Corpus 2025 are structured across three phases to ensure high-quality, actionable data:

\textbf{Phase 1:} Scoring. Each response is holistically and analytically scored. Holistic scoring assigns a CEFR-level grade, while analytic scoring evaluates pronunciation, fluency, task achievement, and other detailed criteria. Only responses scoring A2 or higher are included in the corpus.

\textbf{Phase 2:} Transcription Annotation. Manual transcriptions are generated and refined to capture the learner's spoken output accurately. This includes language errors, hesitations, false starts, and pronunciation errors. A subset of data from Phase 1 will be processed in this phase, and this provides the foundation for error annotation in Phase 3. 

\textbf{Phase 3:} Error Annotation. Transcriptions are further processed to correct grammatical errors, creating fluent versions of the learner’s speech. This supports research in Spoken Grammatical Error Correction (SGEC) and Feedback (SGECF).

The first release of the S\&I Corpus 2025 is divided into training (Train), development (Dev), and evaluation (Eval) sets. The detailed statistics for the subsets are presented in Table~\ref{tab:dev-eval-stats}. Table~\ref{tab:dev-dist} shows the grade distribution for the Dev set (the Train and Eval sets have similar distributions), the majority are in the CEFR range B1-B2+ as can be seen in the Table. Each test part is graded and awarded a corresponding score from 2 (A1) to 5.5 (C1+). The overall score for the four open speaking parts is the average of the scores for each part.
More details about the corpus can be found in~\cite{knill2024sandi}.

\begin{table}[!hbtp]
    \centering
    \begin{tabular}{@{ }p{7mm}|p{5mm}p{6.5mm}|p{5mm}p{5mm}p{5.5mm}|p{5mm}p{4mm}@{}}
    \toprule
       \multirow{2}{*}{Subset} & \multicolumn{2}{c|}{No. of} & \multicolumn{3}{c|}{No. of Hours} & \multicolumn{2}{c}{No. of Words} \\
         & Sub & Utt & Trans & GEC & SLA & Trans &GEC \\
         \midrule
         Train & 6640 & 39490 & 28.2 & 13.0 & 244.2 & 180k & 70k \\ 
         Dev & 438 & 5616 & 22.9 & 20.8 & 35.3 & 140k & 105k \\
         Eval & 442 & 5642 & 22.7 & 20.4 & 35.4 & 140k & 104k \\ 
         \bottomrule
    \end{tabular}
    \caption{Data set statistics for S\&I train, dev and eval sets.}
    \label{tab:dev-eval-stats}
\end{table}

\begin{table}[htbp]
    \centering
    \begin{tabular}{c|c}
    \toprule
       CEFR Grade & \hspace{0.2cm}\% Data \hspace{0.2cm}\\
       \midrule
       A2  & 2.1 \\
       A2+  & 5.7 \\
       B1 & 18.3 \\
       B1+ & 25.3 \\
       B2 & 25.1 \\
       B2+ & 18.3 \\
       C1 & 5.0 \\
       C1+ & 0.2 \\
       \bottomrule
    \end{tabular}
    \caption{Distribution of holistic CEFR grades for Dev set. Eval and Train sets have similar distributions.}
    \label{tab:dev-dist}
\end{table}

\subsection{Text Source for Disfluency Detection}
\label{sec:data_swbd}
The Switchboard Reannotated Dataset\footnote{\label{note5}\url{https://github.com/vickyzayats/switchboard\_corrected\_reannotated}} is an enhanced version of the widely used Switchboard I~\cite{godfrey1992switchboard} and Release 2 corpus~\cite{godfrey1993switchboard}, focusing on the accurate annotation of disfluencies in conversational speech. This dataset aligns the carefully corrected MsState transcriptions~\cite{deshmukh1998resegmentation}, known for their improved word accuracy and time alignments, with disfluencies initially provided in an earlier release of the Switchboard dataset~\cite{marcus1999treebank}. These disfluency annotations capture key speech phenomena such as restart, repetition, and repair, which are critical for studying and modelling spontaneous speech patterns. By combining the precise MsState transcriptions with detailed disfluency labels, the reannotated dataset offers a high-quality resource for developing and evaluating systems in areas such as Disfluency Detection (DD), Automatic Speech Recognition (ASR), and conversational AI. This dataset, including predefined train, dev and test partitions~\cite{charniak2001edit}, is openly available.\footnotemark[\getrefnumber{note5}] This enhanced version of the Switchboard corpus is used to train the baseline Disfluency Detection (DD) system, which is detailed in Section~\ref{sec:sgec}.

\subsection{Text Source for Grammar Error Correction}
\label{sec:data_bea}
The BEA-2019 dataset,\footnote{\url{https://www.cl.cam.ac.uk/research/nl/bea2019st/}} introduced as part of the BEA-2019 Shared Task on Grammatical Error Correction (GEC), is a large collection of English learner texts with annotations for grammatical, lexical, and spelling errors~\cite{bryant2019bea}. It includes data from various subsets, such as the First Certificate in English (FCE), Write \& Improve + LOCNESS, Lang-8 Corpus of Learner English, and National University of Singapore Corpus of Learner English (NUCLE). It includes a diverse range of learner proficiency levels and L1 backgrounds, making it ideal for training and testing GEC systems. We use the BEA-2019 training set to train the baseline GEC system, and the combination of the WI+LOCNESS dev set and the FCE dev set is used for evaluation during training. The data have been stripped of punctuation and capitalisation, while orthographic and spelling errors are retained, as they do not occur in spoken language.

\section{Challenge Introduction}
The Speak \& Improve Challenge 2025 comprises four shared tasks, designed to advance automated language learning assessment and feedback. Two primary tasks, Spoken Language Assessment (\textbf{SLA}) and Spoken Grammatical Error Correction Feedback (\textbf{SGECF}), focus on language learning, with an emphasis on assessing learner proficiency and providing corrective feedback. The other two supplementary tasks, Automatic Speech Recognition (\textbf{ASR}) and Spoken Grammatical Error Correction (\textbf{SGEC}), focus on the technology side, aiming to improve the underlying tools used in speech processing for assessment and learning. Each task presents its own unique challenges, but together they aim to solve critical problems in building comprehensive and fair automated systems for language education. 
 
To ensure a wide range of participation and foster innovation, each task in the challenge includes two tracks:
\begin{itemize}
    \item \textbf{Closed Track:} Participants may only use pre-trained models from the released baseline system, the provided dataset, and named datasets used in training/fine-tuning the models in the baseline system. This track encourages innovation with limited resources and simulates real-world constraints faced in language-learning applications.
    \item \textbf{Open Track:} Participants may use any additional publicly available data and pre-trained models to develop their systems. This track helps explore the upper-bound performance and allows for maximum flexibility in improving system accuracy and robustness.
\end{itemize}

\begin{figure}
    \centering
    \includegraphics[width=0.99\linewidth]{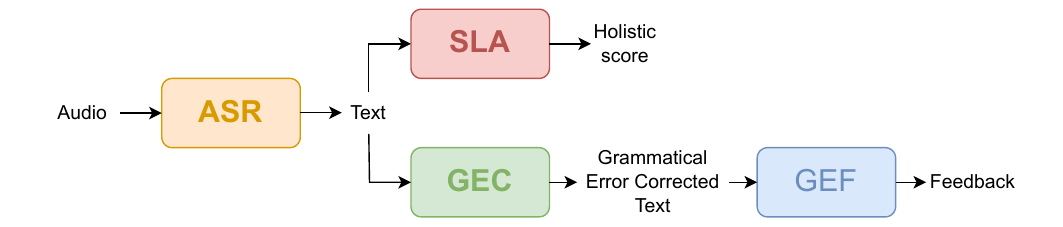}
    \caption{The pipeline of the tasks in the Speak \& Improve Challenge.}
    \label{fig:pipeline}
\end{figure}

A cascaded baseline system and the scoring tool for each task will be supplied for the challenge. As presented in Figure~\ref{fig:pipeline}, the cascaded baseline consists of four components, each representing a task in this challenge.  In the following section detailed descriptions, the baseline system, evaluation and baseline results for each task are presented.

\section{Tasks and Baselines}
\subsection{Automatic Speech Recognition (ASR)}
\label{sec:asr}
The task of Automatic Speech Recognition (ASR) is integral to this challenge, serving as the front-end for all the other tasks. This supplementary task focuses on generating accurate transcriptions of learners' spoken responses, highlighting the importance of advancing the state-of-the-art in ASR within the challenging domain of L2 learners' English speech.
ASR is a foundational technology for many language-learning applications, but recognising speech from learner speech presents significant challenges due to its variability in pronunciation, fluency, and accents. By addressing these complexities, this task encourages participants to develop ASR systems that are more robust and inclusive, offering systems that accommodate learners from diverse linguistic and cultural backgrounds and of a wide range in English speaking proficiency levels.

\textbf{Baseline:} The baseline ASR system for this task will use the OpenAI Whisper small model~\cite{radford2023robust}.

\textbf{Evaluation:} Systems will be evaluated based on Speech Word Error Rate (SpWER)~\cite{ma2023slate}, which is adopted from the Standard WER metric. SpWER aims to assess recognition of what the speaker said - very important for assessment and learning - rather than a standard written transcript so punctuation and case are removed, numbers are written out in orthographic word form, and partial words and hesitations are made optionally deletable in scoring. All forms of hesitations are mapped to a single hesitation token. An evaluation tool based on the NIST sclite scoring tool from SCTK, the NIST Scoring Toolkit\footnote{\href{https://github.com/usnistgov/SCTK/}{https://github.com/usnistgov/SCTK}}
will be provided.

\textbf{Models allowed to be used:} In the Closed Track, participants are restricted to using only OpenAI Whisper models, but any size of the Whisper models is permitted. In the Open Track, participants may use any model structure of their choice.

\textbf{Data sources allowed to be used:} Only the released S\&I data may be used in the Closed Track. Any open-source data may be used in the Open Track.

\textbf{Results:} The baseline ASR result on the S\&I Dev set is presented in Table~\ref{tab:baseline_asr}.

\begin{table}[!htbp]
    \centering
    \begin{tabular}{c|ccc|c}
    \toprule
    Data & Sub & Del & Ins & WER \\
    \midrule
    Dev  & 4.9 & 4.0 & 1.5 & 10.4 \\
    \bottomrule
    \end{tabular}
    \caption{Baseline ASR (\%)WER results on Dev set.}
    \label{tab:baseline_asr}
\end{table}

\subsection{Spoken Language Assessment (SLA)}
\label{sec:sla}
As one of the primary tasks, this task aims to evaluate the holistic proficiency of learners' spoken responses, predicting scores that closely align with human assessments. Participants are encouraged to assess key language features, such as pronunciation, fluency, intonation, and grammatical accuracy. The provided dataset contains holistic scores for each part of the submission as well as for the overall impression of the submission. These scores are rated across various CEFR levels, ranging from A2 (elementary) to C1+ (advanced) as detailed in Section~\ref{sec:data_sandi}. By encompassing a broad range of proficiency levels, the dataset ensures fairness and inclusivity, supporting the development of models that cater to learners with diverse linguistic capabilities.

\textbf{Baseline:} As illustrated in Figure~\ref{fig:pipeline}, the baseline cascaded SLA system comprises an ASR component to convert learners' spoken responses into text and a text grader to predict the score from the text transcription. For each audio response in the dataset, a text transcription generated from the baseline ASR system is provided. Participants can build their SLA systems using either the provided text transcriptions (cascaded system), audio data (end-to-end system), or both (hybrid system). While end-to-end and hybrid systems are viable approaches, a cascaded system is chosen as the baseline for its simplicity and explainability. Specifically, the baseline uses BERT-based text graders, as described in~\cite{raina2020universal, banno2023assessment}. In this system, BERT (\emph{bert-base-uncased}\footnote{\label{note1}\url{https://huggingface.co/google-bert/bert-base-uncased}}) is used to extract word embeddings, which is followed by four multi-head self-attention layers~\cite{vaswani2017attention}. The concatenated 4x768-D embeddings are input to a regression head formed of two fully connected layers, of 600-D and 20-D, respectively. 
ReLU activation is applied to both fully connected layers, and the final layer maps the output to a single value representing the predicted score for each input sequence. Four baseline text graders are trained, each corresponding to a specific part of the test submission, and will predict the holistic score for that part $\hat{pred}_{p}$ (Equation~\ref{eq:sla_part}). These graders are trained on normalised text transcriptions of the respective part in the released S\&I training set, and the text transcriptions were generated from the baseline ASR model (Whisper small). The predictions from each model are averaged to get the overall prediction $\hat{pred}_{o}$, as shown in Equation~\ref{eq:sla_overall}.
\begin{equation}
    \hat{pred}_{p} = SLA(ASR(x))
    \label{eq:sla_part}
\end{equation}
\begin{equation}
    \hat{pred}_{o} = \frac{1}{4} \sum_{p\in {1,3,4,5}} \hat{pred}_{p}
    \label{eq:sla_overall}
\end{equation}


\textbf{Evaluation:} A valid system needs to predict the overall score for a test submission, which consists of four parts. Systems will be evaluated at the test submission level using multiple metrics, including Root Mean Square Error (RMSE), Pearson correlation coefficient (PCC), Spearman's rank coefficient (SRC), and the percentage of the predicted scores that are equal to or lie within 0.5 (i.e., within half a grade) ($\% \leq 0.5$), and within 1.0 (i.e., one grade) ($\% \leq 1.0$) of the actual score. RMSE will be the primary metric for ranking the submissions. An evaluation tool will be provided.

\textbf{Models allowed to be used:} In the Closed Track, only BERT (\emph{bert-base-uncased}\footnotemark[\getrefnumber{note1}]) is allowed to be used. In the Open Track, any publicly available models are allowed.

\textbf{Data sources allowed to be used:} Only the released S\&I data are allowed to be used in the Closed Track. Any open-source data may be used in the Open Track.

\textbf{Results:} The baseline SLA results on S\&I Dev set are presented in Table~\ref{tab:baseline_sla}.
\begin{table}[!htbp]
    \centering
    \begin{tabular}{@{ }c|p{6.5mm}p{6.5mm}p{6.5mm}p{6.5mm}|@{ }cc@{ }}
    \toprule
    Data & RMSE & PCC & SRC & KRC & \%\textless = 0.5 & \%\textless = 1.0 \\
    \midrule
    Dev  &  0.445 & 0.746 & 0.750 & 0.563 & 73.3 & 96.6\\
    \bottomrule
    \end{tabular}
    \caption{Baseline SLA results on Dev set.}
    \label{tab:baseline_sla}
\end{table}

\subsection{Spoken Grammatical Error Correction (SGEC)}
\label{sec:sgec}
This task focuses on Spoken Grammatical Error Correction (SGEC), aiming to identify and correct grammatical errors in spoken language. It serves as a supplementary task to the Spoken Grammatical Error Correction Feedback (SGECF) task. Participants will design systems that can detect and correct grammatical errors, such as tense usage, subject-verb agreement, and sentence structure, using transcriptions of learners' speech. The goal is to produce grammatically corrected transcriptions. 
Spoken grammatical correction presents unique challenges compared to grammatical correction in written text, as it must account for spontaneous speech patterns, disfluencies (i.e., hesitations, repetitions, and false starts), and incomplete sentences commonly found in spoken language. Addressing these challenges will help develop tools that can provide accurate transcriptions to learners from diverse backgrounds, enhancing the robustness and inclusivity of language education tools. 
Participants can develop their SGEC systems based on the audio recordings, ASR transcriptions or both.
For evaluation, manually grammatically corrected transcriptions of fluent manual transcriptions will be used as the reference standard.

\begin{figure}[!htbp]
    \centering
    \includegraphics[width=0.98\linewidth]{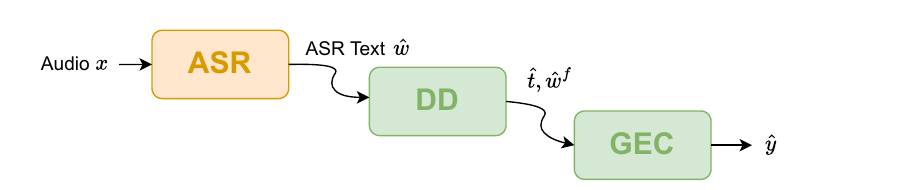}
    \caption{The cascaded SGEC system.}
    \label{fig:sgec}
\end{figure}

\textbf{Baseline:} A cascaded system, as illustrated in Figure~\ref{fig:pipeline}, will be provided as the baseline. It consists of (a) an ASR module which converts the audio sequence $x$ into text $\hat{w}$, (b) a Disfluency Detection (DD) module, which removes disfluencies in the ASR transcriptions, generating reference tags $\hat{t}$ and fluent text $\hat{w}^{f}$, as shown in Equation~\ref{eq:dd}, and (c) a text-based GEC module, which produces the grammatically correct hypotheses $\hat{y}$, as described in Equation~\ref{eq:gec}. This system is similar to the systems described in~\cite{banno2023b_slate_grammatical,banno2024towards}.
\begin{equation}
    \hat{t}, \hat{w}^{f} = DD(ASR(x))
    \label{eq:dd}
\end{equation}
\begin{equation}
    \hat{y} = GEC(DD(ASR(x)))
    \label{eq:gec}
\end{equation}
Specifically, the DD in the baseline is initialised with \emph{bert-base-uncased}\footnotemark[\getrefnumber{note1}] and fine-tuned for binary token classification, i.e., fluent or disfluent tokens. The model is trained with a custom classification head added on top of the BERT model, consisting of a linear layer reducing BERT's 768-dimensional embeddings to 128 dimensions, a dropout layer with a probability of 0.2, and a final linear layer projecting to two output classes. This model is trained on the training set of the Switchboard Reannotated Dataset (described in Section~\ref{sec:data_swbd}) for 3 epochs with a learning rate of 5e-06.
The GEC model is initialised with \emph{bart-large}\footnote{\label{note8}\url{https://huggingface.co/facebook/bart-large}} and fine-tuned on BEA-2019 training set~\ref{sec:data_bea}. The dev set is composed of WI+LOCNESS dev set and FCE dev set. The model is trained with cross-entropy loss with label smoothing set to 0.1 for a total of 8475 iterations (3.75 epochs). The training configuration includes a learning rate of 3e-5, a batch size of 64, gradient accumulation over 4 steps, and a maximum sequence length of 128 tokens. An alignment step is performed to align the GEC transcriptions with the fluent (FLT) transcriptions to get timestamp information for each word. The resulting aligned output is saved in a CTM file. This alignment process will also be included as part of the baseline system release.

\textbf{Evaluation:} Submissions will be evaluated using both Translation Edit Rate (TER) and Word Error Rate (WER), with WER being the main metric for ranking the submissions. These metrics were chosen based on studies in~\cite{lu2022assessing, banno2024towards}, which demonstrate their relevance for spoken GEC. WER is chosen as the primary metric for its simplicity and clarity, making system performance easier to evaluate. Response sentences in the same test submission will be merged, and evaluation will be made at the submission level. A scoring tool with all the necessary text processing steps will be provided.

\textbf{Models allowed to be used:} In the Closed Track, only \emph{bart-large}\footnotemark[\getrefnumber{note8}] (and \emph{bert-base-uncased}\footnotemark[\getrefnumber{note1}] 
if participants want to include DD in the process) is allowed to be used. In the Open Track, any publicly available models are allowed.

\textbf{Data sources allowed to be used:} Only the released S\&I data and the text sources used in the baseline, namely BEA-2019, including WI+LOCNESS dev set and FCE dev set (and optionally the Switchboard Reannotated Dataset if participants want to include DD in the process) are allowed to be used in the Closed Track. Any open-source data can be used in the Open Track.

\textbf{Results:} The baseline SGEC results on the Dev set are presented in Table~\ref{tab:baseline_sgec}.
\begin{table}[!htbp]
    \centering
    \begin{tabular}{c|ccc|c}
    \toprule
    Data & Sub & Del & Ins & WER \\
    \midrule
    Dev  & 8.4 & 4.8 &   4.1  & 17.3 \\
    \bottomrule
    \end{tabular}
    \caption{Baseline SGEC results on Dev set.}
    \label{tab:baseline_sgec}
\end{table}

\subsection{Spoken Grammatical Error Correction Feedback (SGECF)}
This is another primary task in the challenge. In this task, participants will provide spoken grammatical error correction feedback based on the audio and/or ASR generated transcriptions of the audio. The goal is to develop systems that give learners clear, actionable feedback on where and how they have made grammatical errors or have spoken disfluently. Feedback should be easy for learners to understand and apply, offering specific guidance on how to improve their spoken English. While the purpose of SGEC is merely to provide a grammatically correct transcription, SGECF aims to provide feedback information on the types of errors made by learners. By making feedback more accessible and effective, this task helps to enhance the inclusivity of language-learning tools, ensuring they are useful for learners from diverse backgrounds.
Participants will be provided with transcriptions generated by the baseline ASR system, fluent transcriptions with disfluency removed manually, and manually grammatically error corrected transcriptions of fluent manual transcriptions.

\textbf{Baseline:} The baseline for this task is a cascaded system identical to the one provided for the SGEC task. While the baseline system and its outputs remain the same as in the SGEC task, the evaluation reference differs. Participants can develop their SGECF systems based on the audio data, text data or a combination of both.

\textbf{Evaluation:} The MaxMatch (M2)~\cite{dahlmeier2012better} file generated from fluent manual transcriptions (FLT) and manual GEC transcriptions will be used as the reference for evaluation. ERRANT~\cite{bryant2017automatic} F0.5 will be the primary metric for this task.

\textbf{Models and data source allowed to be used:} the same as in the SGEC task.

\textbf{Results:} The baseline SGECF results on the Dev set are presented in Table~\ref{tab:baseline_sgecf}.
\begin{table}[!htbp]
    \centering
    \begin{tabular}{c|ccc}
    \toprule
    Error & P & R & F$_{0.5}$ \\
    \midrule
    R:PREP & 0.643 & 0.172 & 0.416 \\
    M:DET & 0.544 & 0.096 &  0.282 \\
    R:NOUN:NUM & 0.538 & 0.094 & 0.277 \\
    R:VERB:TENSE & 0.258 & 0.069 & 0.166 \\
    R:VERB & 0.143 & 0.034 &  0.087 \\
    U:DET & 0.535 & 0.172 & 0.376 \\
    R:DET & 0.448 & 0.051 & 0.176 \\
    R:NOUN & 0.138 & 0.035 & 0.087 \\
    \midrule
    Overall  &  0.37 & 0.08 & 0.22\\
    \bottomrule
    \end{tabular}
    \caption{Baseline SGECF results on Dev set.}
    \label{tab:baseline_sgecf}
\end{table}

\section{Conclusions and Future Work}
In this paper, we have described the Speak \& Improve Challenge 2025, which aims to advance language learning and technology in spoken language assessment and feedback. The challenge is built upon the Speak \& Improve Corpus 2025 which is being pre-released to support it; a dataset featuring rich annotations and diverse speaker L1 backgrounds. The corpus includes approximately 315 hours of speech data, drawn from open speaking test submissions on the Speak \& Improve platform, and is divided into train, dev, and eval sets.
The challenge features four tasks: ASR, SLA, SGEC and SGECF. For each task, we have introduced a baseline system, described the process of developing it, and provided the evaluation tools. These resources are designed to provide a starting point for participants and set strong benchmarks for innovation.
By providing a comprehensive dataset and clearly defined tasks, we aim to attract a diverse group of participants in the challenge, to make contributions and drive progress in the field of spoken language processing.

\section{Acknowledgements}
This paper reports on research supported by Cambridge University Press \& Assessment (CUP\&A), a department of The Chancellor, Masters, and Scholars of the University of Cambridge. 
The Speak \& Improve Challenge 2025 would not be possible without the data from Speak \& Improve kindly provided by CUP\&A from version 1 of the \href{https://speakandimprove.com}{Speak \& Improve} web application. Many thanks to the team at \href{https://englishlanguageitutoring.com}{English Language iTutoring Ltd (ELiT)} who created and ran the \& Improve learning platform, and the ELiT annotation tool. In particular, Paul Ricketts and Scott Thomas from ELiT helped with establishing and hosting the corpus and with the Challenge set-up. 
Annotations of the corpus are provided by the ELiT humannotator team, with Gloria George and David Barnett providing phase 2 and 3 annotations.
Annabelle Pinnington and her team in \href{https://www.cambridgeenglish.org}{Cambridge University Press \& Assessment's} Multi-Level Testing team have provided all the question prompts for Speak \& Improve. Amanda Cowan and Jing Xu at Cambridge English helped support this Challenge.

\bibliographystyle{IEEEtran}
\bibliography{mybib}

\end{document}